\documentclass{article}
\usepackage{spconf,amsmath,amssymb,graphicx,epsfig,dirtytalk,caption,subcaption,multirow,float,array}

\def\Sup{{\cal S}}
\def\UnsupA{{\cal U_A}}
\def\UnsupT{{\cal U_T}}
\DeclareMathOperator*{\argmax}{arg\,max}

\title{DUAL LEARNING FOR LARGE VOCABULARY ON-DEVICE ASR}
\name{Cal Peyser$^{12}$, Ronny Huang$^2$, Tara Sainath$^2$, Rohit Prabhavalkar$^2$, Michael Picheny$^1$, Kyunghyun Cho$^1$}
\address{
    $^1$Center for Data Science, New York University, New York City, USA \\
    $^2$Google Inc., U.S.A}
\copyrightnotice{978-1-6654-7189-3/22/\$31.00~\copyright2023 IEEE}
\begin{document}
%
\maketitle
\begin{abstract}
Dual learning is a paradigm for semi-supervised machine learning that seeks to leverage unsupervised data by solving two opposite tasks at once.  In this scheme, each model is used to generate pseudo-labels for unlabeled examples that are used to train the other model.  Dual learning has seen some use in speech processing by pairing ASR and TTS as dual tasks.  However, these results mostly address only the case of using unpaired examples to compensate for very small supervised datasets, and mostly on large, non-streaming models.  Dual learning has not yet been proven effective for using unsupervised data to improve realistic on-device streaming models that are already trained on large supervised corpora.  We provide this missing piece though an analysis of an on-device-sized streaming conformer trained on the entirety of Librispeech, showing relative WER improvements of $10.7\%/5.2\%$ without an LM and $11.7\%/16.4\%$ with an LM.  
\end{abstract}
%
%
\section{INTRODUCTION}
\label{sec:intro}

The high cost of supervision in speech datasets has yielded a rich literature on the use of unpaired audio and text to improve ASR performance.  The conventional setting of this problem involves use of unpaired data to enable a working ASR system to be trained on a small enough amount of paired data that ASR would otherwise have been unfeasible.  Work on this setting of the problem dates back to long before deep neural networks became the dominant approach to speech processing \cite{Nguyen2004, Lamel2002, Ma2008}.

In the last several years, progress on this problem has advanced to the point where reasonably strong ASR systems can be trained with only tens of hours of supervised data.  Methods for doing so fall broadly into two categories: \emph{generative} methods that model the data distribution itself, and \emph{contrastive} methods that model the likelihood of a sample given some context. 

Generative methods are premised originally on the autoencoder \cite{Kaiser2013}, which when applied in the speech domain has been shown to yield strong representations \cite{Chorowski2019}.  \say{Predictive coding} extends the autoencoder by seeking to predict a sample some number of steps in the future, and can be done autoregressively \cite{Chung2020} or non-autoregressively \cite{Liu2021}.  The success of masked language modeling in the language domain (e.g. \cite{BERT}) motivated the application of masked prediction \cite{Jiang2020, Liu2020Mockingjay} and discretization \cite{DiscreteBERT} to generative speech pretraining, yielding strong results.

Contrastive methods, on the other hand, are premised on \say{contrastive predictive coding} (CPC) \cite{CPC}, in which positive examples and distractors are together used to model how much more or less likely an audio sample is given some context.  In principle, this relieves the system from modeling irrelevant details of the data distribution.  The Wav2Vec line of papers \cite{Wav2Vec, VQVav2Vec, Wav2Vec2} demonstrated the viability of CPC features for ASR, eventually combining CPC with generative methods to achieve strong performance on Librispeech using as little as ten minutes of supervised audio.  WavLM \cite{WavLM} iterated on these ideas and demonstrated usefulness on an array of speech tasks beyond ASR.  

While these ideas have enjoyed great success, applications have strongly emphasized the low resource setting and generally involve very large, full-context models.  That is, while we've seen that we can use unlabeled speech to build an ASR model given very little supervised data, it is not yet clear that we can use it to overcome the challenges inherent in on-device systems on high resource languages.  We offer an alternative framing of the problem, in which a large amount of supervised data is available, but the ASR system must operate under the real-world constraints of being able to stream results and being sufficiently small so as to fit on a smartphone.  We ask how unsupervised audio and text may be used to improve performance in this setting. 

Dual learning is a method in semi-supervised learning in which two opposite tasks are learned simultaneously, with each model providing supervision for the other.  This idea has been successful in machine translation in the form of \say{backtranslation} \cite{He2016, Hoang2018}.  In the speech domain, TTS has been used as the \say{dual} task to ASR and combined with audio and text reconstruction tasks to some success \cite{Tjandra2017, Ren2019, Xu2020, CycleCon}.  However, these results are constrained to the low-resource setting, with non-streaming models and usually with short utterance lengths.  The TTS4ASR line of work \cite{Tts4Asr, Tts4Pretrain, Tts4Pretrain2} uses an approach similar to dual learning, in which a TTS model is used to provide supervision for unpaired audio before training.  These models have been proven to work with large supervised datasets; \cite{Tts4Pretrain2} in particular yields gains on top of a strong Librispeech baseline.  However, even these models do not address the on-device setting, as they are very large (\cite{Tts4Pretrain2} has 600M parameters) and are non-streaming.

In this paper, we seek to fill what we see is a missing piece in the literature and demonstrate the viability of dual learning in the large vocabulary, on-device setting.  We show consistent improvements over a supervised streaming conformer optimized for on-device inference trained on all 960 hours of Librispeech.

The rest of this paper is organized as follows.  Section \ref{sec:methods} outlines our model and training procedure and specifies choices that are necessary to scale the method.  Section \ref{sec:experiments} details the setup of our experiments.  Section \ref{sec:results} presents our results and analysis and we conclude in Section \ref{sec:conclusion}.

\section{METHODS}
\label{sec:methods}
In this section, we describe our implementation of ASR pretraining based on dual learning.  

\subsection{ARCHITECTURE}
\label{ssec:methods_arch}
In order to perform ASR, TTS, and reconstruction in both domains,  our implementation must include encoders and decoders for both audio and text.  Imitating \cite{CascadeConformer}, we implement streaming with an architecture that can emit a provisional hypothesis immediately and then revise it after a short delay. In order to improve the likelihood of success on the two more difficult tasks (ASR and TTS), we adapt these components from existing ASR and TTS architectures.  Our audio encoders and text decoder are adapted from conformer \cite{Conformer, CascadeConformer}.  Our text encoder and audio decoder are adapted from Tacotron 2 \cite{Tacotron2}.

\begin{figure}[H]
\centering
\includegraphics[scale=0.25]{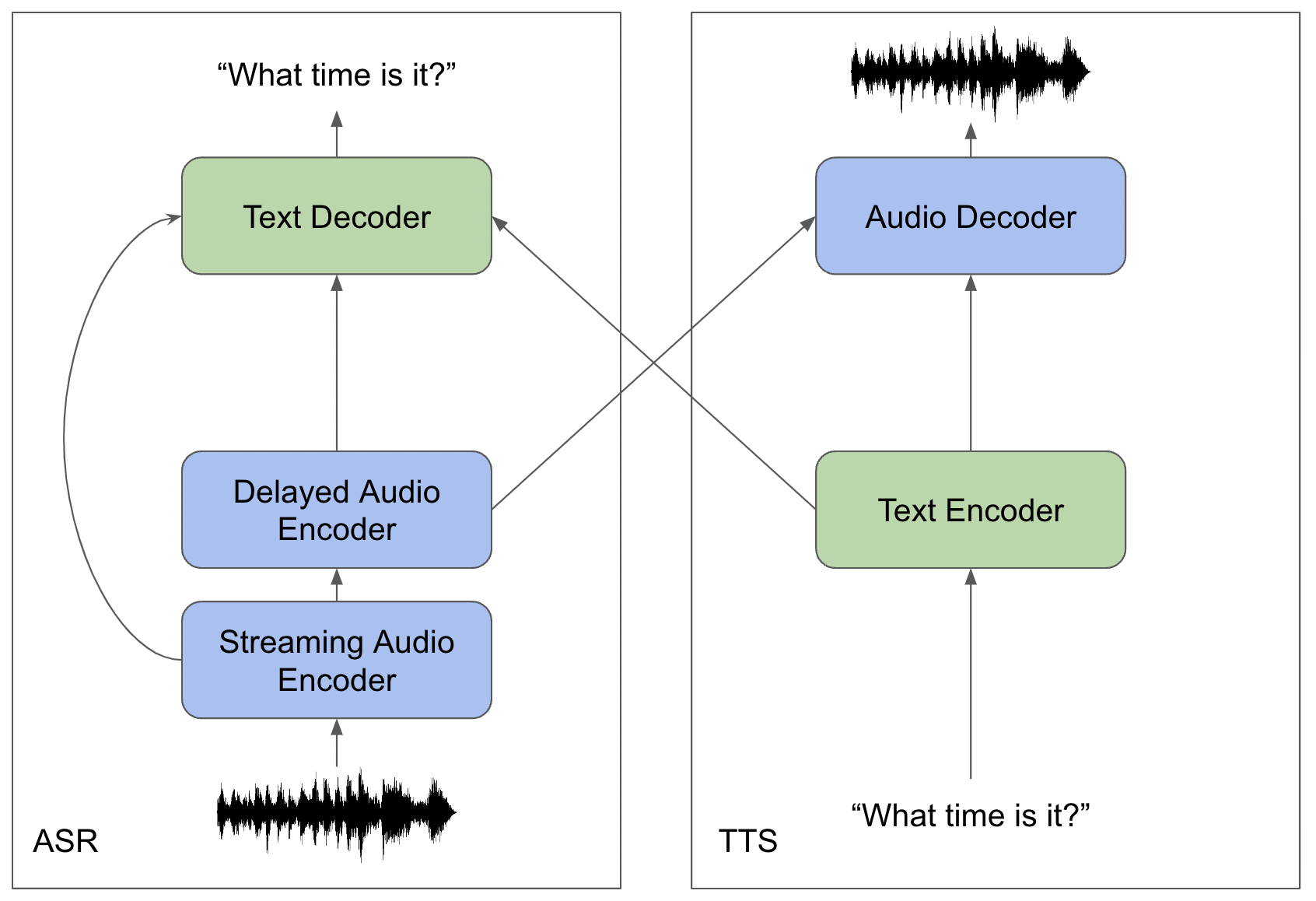}
\caption{The architecture of our dual learning model.  Blue components participate in audio reconstruction, while green components participate in text reconstruction.}
\label{fig:cascade}
\vspace{-0.1in}
\end{figure}

Formally, we frame our problem around three datasets. First, $\Sup$ consists of paired text and audio examples $(x, y)$, where $x = (x_0, ...,x_m)$ is an audio sample of length $m$ and $y = (y_0,...,y_n)$ is the corresponding text transcript of length $n$.  Second, $\UnsupT$ gives unpaired text examples $y$, and finally $\UnsupA$ gives unpaired audio examples $x$.  We then define the components of the model as functions. We define the audio encoders $E_A^\text{streaming}$, which has only left-context and $E_A^\text{delay}$, which has 900ms of right-context.  We also define the text encoder $E_T$, the decoders $D_A$ and $D_T$, and the linear transformations $T_{A\rightarrow T}$ which maps an audio embedding to a text embedding and $T_{A\leftarrow T}$ which maps a text embedding to an audio embedding.

We may then proceed to define the model's objectives.

\subsubsection{Supervised ASR and TTS}
We follow \cite{CascadeConformer} to build a model capable of emitting an ASR hypothesis in real time while streaming finalized predictions with 900ms latency.  To this end, we split ASR training into two losses.  For $(x, y) \in \Sup$, the immediate streaming task is given by:

\[\mathcal{L}_{\text{ASR}}^{\text{streaming}} = L_{\text{xent}}(y, D_T \circ E_A^{\text{streaming}}(x)) \]
\\
while the delayed task is given by:
\\
\[ \mathcal{L}_{\text{ASR}}^{\text{delay}} = L_{\text{xent}}(y, D_T \circ E_A^{\text{delay}} \circ E_A^{\text{streaming}}(x)) \]

where $\circ$ denotes function composition. That is, the streaming task uses only the first, left-context encoder while the delayed task adds a further encoder with 900ms of right-context.  Here we have defined $L_{\text{xent}}$ as the cross-entropy loss over text units.

Since this work focuses on pretraining for ASR systems, we do not seek to stream the TTS task.  Instead we define the single full-context task:
\[ \mathcal{L}_{\text{TTS}} = L_{\text{MSE}}(x, D_A \circ E_T(y)) \]
where we have defined $L_{\text{MSE}}$ as the mean squared error loss over continuous audio features.

\subsubsection{Unsupervised ASR and TTS}
Dual learning for speech and text involves the use of the TTS system to provide pseudo-labels for the ASR system and visa versa.  Specifically, for an unpaired text example $y \in \UnsupT$ we derive the pseudo-label $\hat{y}$  by beam search over the outputs of the ASR model.  We may then define the objectives:

\[\mathcal{L}_{\text{U-ASR}}^{\text{streaming}} = L_{\text{xent}}(y, D_T \circ E_A^{\text{streaming}}(\hat{y})) \]
\\
and
\\
\[ \mathcal{L}_{\text{U-ASR}}^{\text{delay}} = L_{\text{xent}}(y, D_T \circ E_A^{\text{delay}} \circ E_A^{\text{streaming}}(\hat{y})) \]
\\
Similarly, for an unpaired audio example $x \in \UnsupA$ we may derive the pseudo-label $\hat{x}$ by performing inference in the TTS system.  We may then define the objective:

\[ \mathcal{L}_{\text{U-TTS}} = L_{\text{MSE}}(x, D_A \circ E_T(\hat{x})) \]

\subsubsection{Reconstruction}
To perform text reconstruction, we must pass representations from the ASR encoder to the TTS decoder and visa versa.  We find that when initializing using pre-trained ASR and TTS systems, the components struggle to adapt to each-other and the model fails to converge.  We find that this problem is alleviated simply by placing a single linear transformation $T_{A\rightarrow T}$ between the audio encoder and audio decoder, and another transformation $T_{A\leftarrow T}$ between the text encoder and text decoder.  With this in mind, we define the text reconstruction task:

\[ \mathcal{L}_{\text{Text Recon}} = L_{\text{MSE}}(y, D_T \circ T_{A\leftarrow T} \circ E_T)(y) \]
for $(x, y) \in \Sup$, with $\mathcal{L}_{\text{U-Text Recon}}$ defined analogously for $x \in \UnsupT$. We similarly define the audio reconstruction task:

\[ \mathcal{L}_{\text{Audio Recon}} = L_{\text{xent}}(x, D_A \circ T_{A\rightarrow T} * E_A^{\text{delay}} \circ E_A^{\text{streaming}}(x)) \]
for $(x, y) \in \Sup$, with $\mathcal{L}_{\text{U-Audio Recon}}$ defined analogously for $x \in \UnsupA$.

\subsection{Training}
\label{ssec:methods_training}
We might naively seek to train the above tasks together by alternating tasks across sequences of batches.  We find that such a training scheme fails to achieve convergence, as each task is forgotten during the training of the others.  Instead, we combine all the above tasks in a single batch.  Each batch is split in thirds, the first coming from $\Sup$, the second from $\UnsupA$, and the last from $\UnsupT$.  For the first third, we jointly optimize the supervised tasks:
\[\mathcal{L}_S = \frac{\mathcal{L}_\text{ASR}^{\text{streaming}} + \mathcal{L}_\text{ASR}^{\text{delay}}}{2} + \mathcal{L}_{TTS} + \mathcal{L}_{\text{Text Recon}} + \mathcal{L}_{\text{Audio Recon}}\]
for the second third, we optimize the unsupervised audio tasks:
\[\mathcal{L}_{A} = \mathcal{L}_{\text{U-TTS}} + \mathcal{L}_{\text{U-Audio Recon}}\]
for the last third, we optimize the unsupervised text tasks:
\[\mathcal{L}_S = \frac{\mathcal{L}_\text{U-ASR}^{\text{streaming}} + \mathcal{L}_\text{U-ASR}^{\text{delay}}}{2} + \mathcal{L}_{\text{U-Text Recon}} \]
\\ \\
We find that this method achieves convergence, so long as we initialize the model's components from an ASR and TTS system trained on $\Sup$.  Otherwise, the model generates incorrect pseudo-labels early in training, preventing progress.

\subsection{Language Modeling}
\label{ssec:methods_training}
Since dual learning involves the incorporation of unpaired text at training time, we naturally want to compare our method to the incorporation of unpaired text at inference time.  To this end we evaluate our models with shallow fusion \cite{ShallowFusion} with a pretrained LM.  We also use a Hybrid Autoregressive Transducer (HAT) \cite{HAT} text decoder, which permits the factorization of our models' internal LM.  Ultimately, at inference time for audio sample $x$ we seek:

\[y^* = \argmax_y \log P(y | x) + \alpha P_{ELM}(y) - \beta P_{ILM}(y)  \]
where $P(y|x)$ gives our acoustic model posterior, $P_{ELM}(y)$ gives the likelihood of a transcript in the external LM, $P_{ILM}(y)$ gives the likelihood of the transcript in the internal LM (as formulated in HAT), and $\alpha$ and $\beta$ are hyperparameters.

\begin{table*}[t]
\begin{subtable}{0.49\textwidth}
\centering
\begin{tabular}{||m{6em}|m{4em} m{4em} m{4em}|} 
 \hline
 Model & Baseline & Shallow Fusion & Internal LM \\ [0.5ex] 
 \hline\hline
 \textbf{BASELINE} & 8.4 & 6.3 & 5.8 \\ 
 \hline
 \textbf{E-ALL} & 7.5 & 5.6 & 5.5 \\
 \hline
 \textbf{E-DL} & 8.1 & 6.2 & 6  \\
 \hline
 \textbf{E-RECON} & 10.3 & 7.4 & 7.2\\
 \hline
\end{tabular}
\centering
\caption{Test Clean}
\end{subtable}
\begin{subtable}{0.49\textwidth}
\centering
\begin{tabular}{||m{6em}|m{4em} m{4em} m{4em}|} 
 \hline
 Model & Baseline & Shallow Fusion & Internal LM \\ [0.5ex] 
 \hline\hline
 \textbf{BASELINE} & 22.9 & 19.5 & 18.3 \\ 
 \hline
 \textbf{E-ALL} & 20.4 & 16.3 & 16.2 \\
 \hline
 \textbf{E-DL} & 21.9 & 18.3 & 17.9  \\
 \hline
 \textbf{E-RECON} & 27 & 22.5 & 21.9 \\
 \hline
\end{tabular}
\centering
\caption{Test Other}
\end{subtable}
\caption{WER percentage results on the Librispeech test sets. Baseline evals include no language model.  Shallow Fusion evaluations include LM interpolation with $\alpha=0.2$.  Internal LM evaluations further subtract out the internal LM with $\beta=0.1$.}
\label{table:results}
\end{table*}

\section{EXPERIMENTAL SETUP}
\label{sec:experiments}
In this section, we give the details of our experimental setup.

\subsection{Model}
\label{ssec:experiments_model}
The ASR branch of our model is a cascading conformer adapted from \cite{CascadeConformer}, specifically sized to be realistic for an on-device streaming application.  The streaming encoder is small to ensure fast inference.  It consists first of 3 convolutional layers followed by 7 conformer layers with a 512-dimensional representation for a total of 56M parameters.  The delayed encoder is larger, and is parameterized by 10 conformer layers with a 640-dimensional representation for a total of 99M parameters.  Following \cite{BotrosSainath}, the HAT decoder consists of an embedding network and joint network, contributing another 9M parameters. 

The TTS branch of our model is adapted from Tacotron 2 \cite{Tacotron2}.  The text encoder first maps wordpieces into a 512-dimensional embedding space, followed by three convolutional layers and a single bidirectional LSTM layer, totalling 8M parameters.  The audio decoder consumes the audio sample autoregressively through a \say{pre-net}, which consists of two fully-connected layers with 50\% dropout at each layer.  We find that this aggressive dropout is critical to convergence, since during multi-task training with teacher forcing the TTS decoder has a strong tendency to rely entirely on the autoregressive signal instead of the encoder representation.  This yields poor performance at inference, which in turn creates poor pseudo-labels for unpaired text.  Scheduled sampling \cite{ScheduledSampling} was investigated as an alternative, but was found to be less effective than simple dropout.  The audio sample is then passed to two LSTM layers, which also consume the encoder representation via cross-attention.  After the LSTM has generated an audio prediction, that result is further processed by a full-context \say{post-net} consisting of five convolutional layers.  In total, the TTS decoder has about 26M parameters.

\subsection{Data}
\label{ssec:experiments_data}
We use 960 hours of supervised audio from Librispeech as $\Sup$, 60k hours of unsupervised audio from Librilight \cite{Librilight} as $\UnsupA$, and 80M transcripts from the Librispeech LM set as $\UnsupT$.  We process the audio into a 128-dimensional log-mel feature per 10ms of audio.  We stack every third such feature with the three features before it, yielding 512-dimensional features at 30ms intervals.  We then apply SpecAugment \cite{SpecAugment} with mask parameter $F=27$ and ten time masks, as in \cite{Conformer}.  This forms the inputs to the audio encoder.
 
\subsection{Evaluation}
\label{ssec:experiments_eavl}
We evaluate our models using a beam search with a beam size of 8.  For fusion experiments, we use an external language model trained on $\UnsupT$.  The LM is a causal transformer \cite{Transformer} with 8 layers, 16 attention heads, and a model dimension of 1024, totaling about 100M parameters.

\section{RESULTS}
\label{sec:results}
We evaluate our model relative to a baseline ASR system trained on 960 hours of Librispeech (\textbf{BASELINE}).  The difference between our baseline WER and those reported in full-context works like \cite{WavLM} reflect the added difficulty of streaming results as well as the reduced model size.  We contrast this with a model trained on all tasks defined above (\textbf{E-ALL}). We perform ablations by also training a model on only the supervised and dual learning tasks (\textbf{E-DL}, excluding $\mathcal{L}_\text{Text Recon}$, $\mathcal{L}_\text{U-Text Recon}$, $\mathcal{L}_\text{Audio Recon}$, and $\mathcal{L}_\text{U-Audio Recon}$) and another only on the supervised and reconstruction tasks (\textbf{E-RECON}, excluding $\mathcal{L}_{\text{U-ASR}}^{\text{streaming}}$, $\mathcal{L}_{\text{U-ASR}}^{\text{delay}}$, and  $\mathcal{L}_{\text{U-TTS}}$).  Results are given in Table \ref{table:results}.

We find our method to improve performance on the test-clean/test-other test sets by $10.7\%/5.2\%$ without an LM and $11.1\%/16.4\%$ with an LM included via shallow fusion.  Interestingly, we find that while dual learning alone $(\textbf{E-DL})$ yields improvements, reconstruction alone ($\textbf{E-RECON}$) does not.  Nevertheless, the combination of dual learning and reconstruction ($\textbf{E-ALL}$) yields better results than either alone.  This suggests that reconstruction itself distracts from the ASR tasks but synergizes with dual learning.  This may reflect the fact that on the multi-speaker, long utterances of Librispeech, a joint model benefits from extra exposure to the unsupervised training data in order to produce strong pseduo-labels for dual learning.  

\subsection{Effect of the Language Model}
\begin{figure*}[t]
  \centering    
  \begin{subfigure}[t]{0.5\columnwidth}
	  \includegraphics[width=1.0\columnwidth]{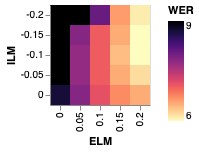}
	  \caption{Baseline, Test Clean}
  \end{subfigure}
  \begin{subfigure}[t]{0.5\columnwidth}  
	  \includegraphics[width=1.0\columnwidth]{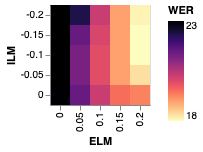}
	  \caption{Baseline, Test Other}
  \end{subfigure}
  \begin{subfigure}[t]{0.5\columnwidth}
  	\includegraphics[width=1.0\columnwidth]{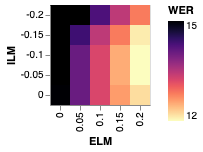}
	\caption{Baseline, LM Only}
  \end{subfigure}
  
  \begin{subfigure}[t]{0.5\columnwidth}
  	\includegraphics[width=1.0\columnwidth]{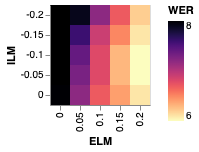}
	\caption{Dual Learning, Test Clean}
  \end{subfigure}
  \begin{subfigure}[t]{0.5\columnwidth}
	  \includegraphics[width=1.0\columnwidth]{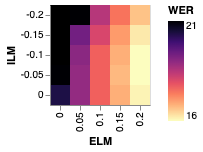}
	  \caption{Dual Learning, Test Other}
  \end{subfigure}
  \begin{subfigure}[t]{0.5\columnwidth}
	  \includegraphics[width=1.0\columnwidth]{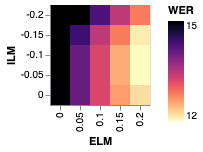}
	  \caption{Dual Learning, LM Only}
  \end{subfigure}
  \caption{The effects of external LM (ELM) and internal LM (ILM) interpolation at inference time.}
  \label{fig:heatmaps}
\end{figure*}

We note with interest that while the application of shallow fusion preserves the gains yielded by our method, further subtracting out the internal language model via HAT only partially preserves those gains.  That is, subtracting out the internal language model substantially closes the gap between the baseline and our method.  Figure \ref{fig:heatmaps} illustrates this effect by plotting WER for a parameter sweep of external LM interpolation (shallow fusion) weights and internal LM interpolation (HAT) weights.  Quantitatively, subtracting out the internal LM with a factor of $\beta=0.1$ from a model with shallow fusion improves $\textbf{BASELINE}$ by $7.9\%/6.2\%$ while only improving $\textbf{E-ALL}$ by $1.2\%/0.6\%$.

This result suggests that our method largely benefits the internal language representation of the ASR system's decoder.  This explanation is consistent with other methods of incorporating unsupervised text into a model at training time such as deep fusion \cite{DeepFusion} and cold fusion \cite{ColdFusion}, but without adding any additional parameters to the ASR model at inference time.  That is, unlike conventional language model fusion, our method bakes knowledge of that data into the parameters of the decoder, providing much the same effect as combining the ASR system with a pretrained language model with no modifications to the architecture.

This result also suggests that the improvements due to our method come mostly, but not entirely, from the unsupervised text data, as opposed to the unsupervised audio.  This is consistent with our design; unsupervised text yields pseudo-labeled examples for the ASR task, while unsupervised audio yields pseudo-labeled examples for the TTS task.

\subsection{Tail Analysis}
Since unsupervised data is often used to address parts of the data distribution that are absent from the supervised training set, we seek to understand the effect of our method on \say{tail} words, which we define as words that are underrepresented in the training data relative to their frequency in the language as a whole.  To this end, we draw inspiration from \cite{LMSets} and generate a test set of examples containing words that are not represented in $\Sup$ but are represented in $\UnsupT$.  Specifically, for some parameter $0 < \tau < 1$ we sample transcripts from $\UnsupT$ containing at least one unigram that occurs with frequency less than $\tau$ in $\Sup$ but greater than $\tau$ in $\UnsupT$.  We set $\tau = 0.00001$ and generate a test set of 10k transcripts.  We then synthesize audio transcripts using a Tacotron TTS system as in \cite{Tacotron2}.  Results on this tail test set are given in Table \ref{table:results}, and measurements of internal and external LM integration are included in Figure \ref{fig:heatmaps}.

\begin{table}[h]
\begin{subtable}{0.5\textwidth}
\centering
\begin{tabular}{||m{6em}|m{4em} m{4em} m{4em}|} 
 \hline
 Model & Baseline & Shallow Fusion & Internal LM \\ [0.5ex] 
 \hline\hline
 \textbf{BASELINE} & 16.1 & 13.8 & 12.9 \\ 
 \hline
 \textbf{E-ALL} & 15.3 & 12.6 & 12.4 \\
 \hline
 \textbf{E-DL} & 15.5 & 13.1 & 12.8  \\
 \hline
 \textbf{E-RECON} & 17.4 & 14.5 & 14.2 \\
 \hline
\end{tabular}
\centering
\end{subtable}
\caption{WER percentage results on the synthesized Tail test set.}
\label{table:lm_only_results}
\end{table}

As above, we find that our method yields improvements across the board.  However, we note that the improvements are smaller than they are on Librispeech test sets.  In particular, without an LM we improve by $5.0\%$, with shallow fusion we improve by $8.7\%$, and with internal language model subtraction we improve by $3.9\%$.

\section{CONCLUSION}
\label{sec:conclusion}
In this paper, we demonstrate that dual learning is an effective method for pretraining an on-device, streaming ASR model using both unsupervised audio and text data.  We provide an analysis that suggests that the majority of improvements due to the method are attributable to a refinement of the ASR decoder's internal language representation.  We believe that this work motivates exploration in how proven methods for modeling unsupervised audio such as discretization and masked language modeling could further improve these results.

	

\bibliographystyle{IEEEbib}
\bibliography{main}

\end{document}